\documentclass[10pt,twocolumn,letterpaper]{article}

\usepackage[pagenumbers]{cvpr} %

\usepackage{graphicx}
\usepackage{amsmath}
\usepackage{amssymb}
\usepackage{booktabs}

\usepackage[pagebackref,breaklinks,colorlinks]{hyperref}

\usepackage[capitalize]{cleveref}
\crefname{section}{Sec.}{Secs.}
\Crefname{section}{Section}{Sections}
\Crefname{table}{Table}{Tables}
\crefname{table}{Tab.}{Tabs.}

\DeclareMathOperator*{\argmin}{arg\,min}
\def\cvprPaperID{7019} %
\def\confName{CVPR}
\def\confYear{2022}

\begin{document}

\title{iLabel: Interactive Neural Scene Labelling}

\author{
Shuaifeng Zhi$^{1}$\thanks{Authors contributed equally to this work.} \quad Edgar Sucar$^{1}$\footnotemark[1] \quad Andre Mouton$^{2}$ \quad Iain Haughton$^{2}$ \\ Tristan Laidlow$^{1}$ \quad Andrew J. Davison$^{1}$\\
$^{1}$ Dyson Robotics Lab, Imperial College\\ 
$^{2}$ Dyson Ltd.\\
{\tt\small \{s.zhi17,e.sucar18\}@imperial.ac.uk}
}
\maketitle

\begin{abstract}

Joint representation of geometry, colour and semantics using a 3D neural field enables accurate dense labelling from ultra-sparse interactions as a user reconstructs a  scene in real-time using a handheld RGB-D sensor. Our iLabel system requires no training data, yet can densely label scenes more accurately than standard methods trained on large, expensively labelled image datasets. Furthermore, it works in an `open set' manner, with semantic classes defined on the fly by the user.

iLabel's underlying model is a multilayer perceptron (MLP) trained from scratch in real-time to learn a joint  neural scene representation.
The scene model is updated and visualised in real-time, allowing the user to focus interactions to achieve efficient labelling.  A room or similar scene can be accurately labelled into 10+ semantic categories with only a few tens of clicks. Quantitative labelling accuracy scales powerfully with the number of clicks, and rapidly surpasses standard pre-trained semantic segmentation methods. We also demonstrate a hierarchical labelling variant. 
\end{abstract}

\vspace{-3.5mm}
\section{Introduction}
\label{sec:intro}

An intelligent agent must build an internal representation of its environment which goes beyond geometry and colour to include a semantic understanding of the scene. Research on neural field representations has shown that an MLP network can be trained from scratch in a single scene via automatic self-supervision to accurately and flexibly represent geometry and appearance \cite{Mildenhall:etal:ECCV2020, Sucar:etal:ICCV2021}. In this paper we demonstrate that the internal scene structure learned by the network allows for efficient user-guided scene segmentation.

\begin{figure}[t]
    \centering
    \includegraphics[width =0.9\linewidth]{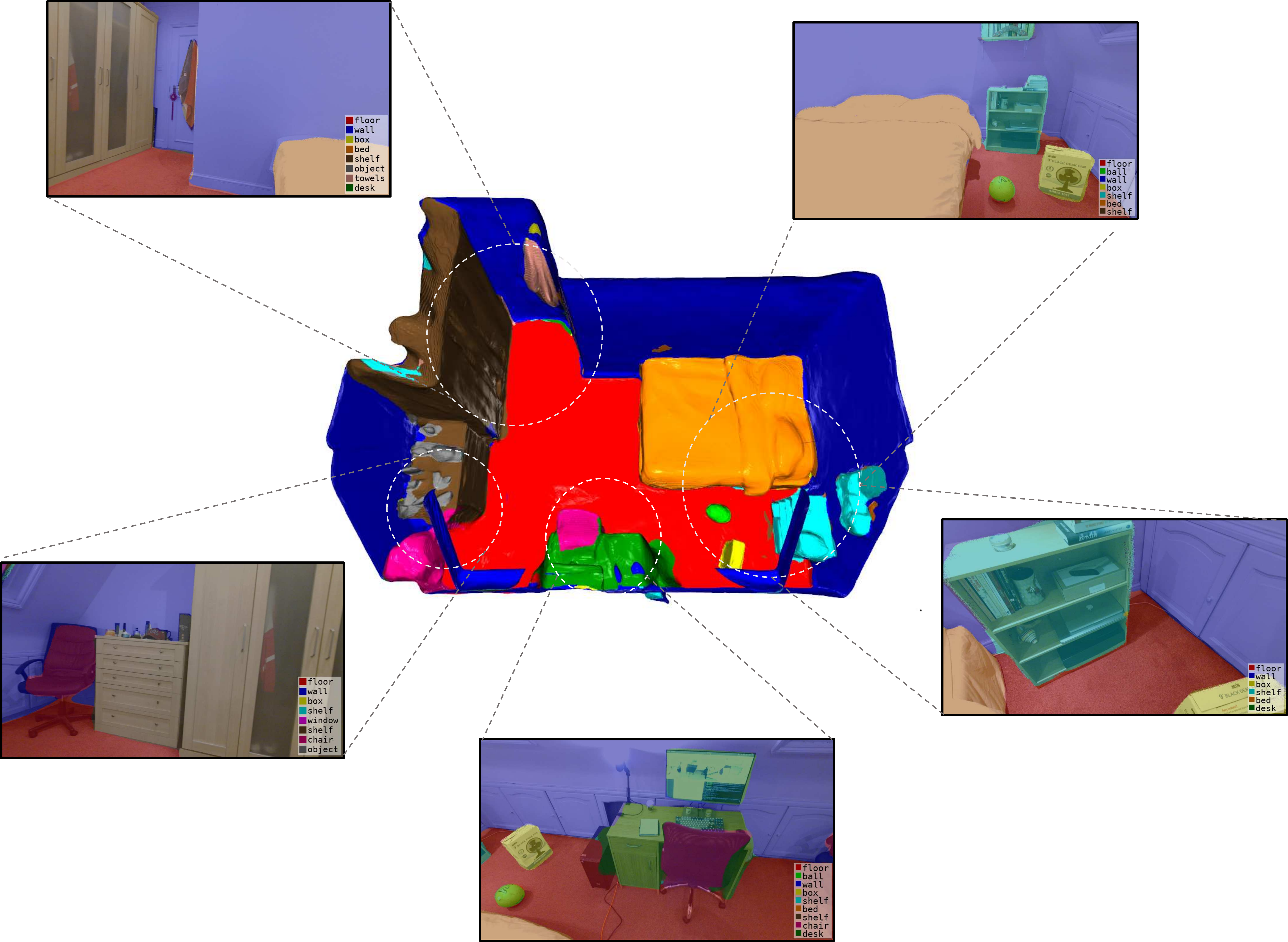}
    \caption{Whole-room semantic mesh labelled in real-time from only 140 interactive clicks and no prior training data. See \url{https://youtu.be/bL7RZaMhRbk} for a video demonstration.}
    \label{fig:room_seg}
        \vspace{2mm} \hrule
\end{figure}

We introduce iLabel, the first online and interactive 3D scene capturing system with a unified neural field representation, which allows a user to achieve high-quality, dense scene reconstruction and multi-class semantic labelling from scratch with only minutes of scanning and a few tens of semantic click annotations. A real-time neural field SLAM forms the basis of our system. The user simultaneously scans a scene and provides sparse semantic annotations on selected keyframe images. By supervising the network on the sparse annotations, semantics are automatically propagated.  The ability to render full predictions in real-time allows a user-in-the-loop to place annotations efficiently, fixing incorrect predictions or adding new classes.

Our approach requires no prior training on semantic datasets, and can therefore be applied in novel contexts, with categories defined on-the-fly by the user in an open-set manner. Standard methods for semantic scene segmentation use deep networks trained on datasets of thousands of images with dense, high-quality human annotations; even then they often have poor performance when the test scene is not a good match for the training set. We show that the quantitative labelling accuracy of iLabel scales powerfully with the number of clicks, and rapidly surpasses the accuracy of standard pre-trained semantic segmentation methods.

Alongside our core iLabel system for multi-class scene labelling via clicks, we present two variations. First, we show that hierarchical semantic labelling can be achieved by interpreting outputs as branches in a binary tree. Second, we demonstrate a `hands free' labelling mode where an automatic uncertainty-guided framework selects a sequence of pixels for which to ask the user for label names without the need for clicks.

The only comparable interactive scene understanding system is SemanticPaint \cite{Valentin:etal:ACMTOG2015}, which trains a classifier on top of a separate dense SLAM system. It requires alternating between training and prediction modes, making labelling cumbersome. We argue that the unified scene representation in iLabel is simpler and more user friendly, and also show qualitatively that iLabel obtains much more precise and complete segmentations.

We demonstrate iLabel in a wide variety of environments, from tabletop scenes to entire rooms and even outdoors. We believe iLabel to be a powerful and user-friendly tool, with much potential for interactive scene labelling for augmented reality or robotics, as well as providing intuitive insights into the ability of neural fields to jointly represent correlated quantities.

\section{Related Work}
\label{sec:related}
\begin{figure*}[!t]
    \centering
        \includegraphics[width =1\linewidth]{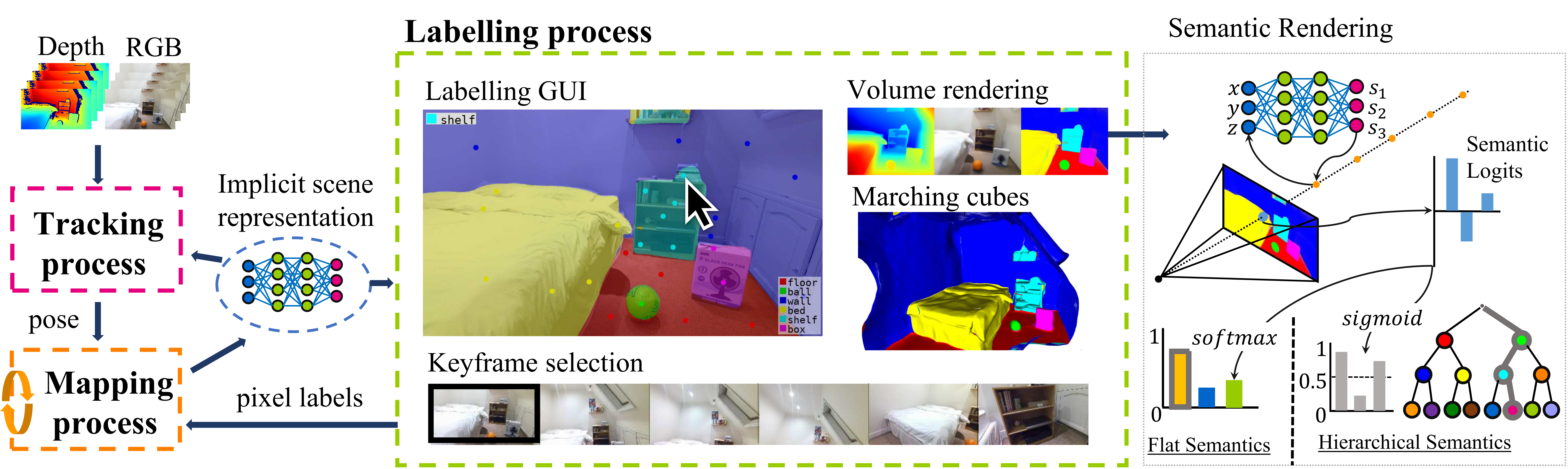}
    \caption{Overview of the iLabel system pipeline.}
    \label{fig:overview}
        \vspace{2mm} \hrule
\end{figure*}
\subsection{iLabel System Overview}

\vspace{-1mm}
\paragraph{Scene Representation for Visual SLAM}
Scene representation in visual SLAM has gradually progressed from sparse feature point sets \cite{Davison:ICCV2003,Engel:etal:ECCV2014,Mur-Artal:etal:TRO2017} 
to dense geometric 3D maps (e.g. surfels, meshes and voxels) \cite{Newcombe:etal:ISMAR2011,Whelan:etal:IJRR2016,Niessner:etal:SIGGRAPH2013,Dai:etal:ACMTOG2017}
and more recently, to neural representations \cite{Bloesch:etal:CVPR2018}, increasingly involving semantics
\cite{Salas-Moreno:etal:CVPR2013,McCormac:etal:ICRA2017,Sunderhauf:etal:IROS2017,Narita:etal:IROS2019,Zhi:etal:CVPR2019,Zhi:etal:ICCV2021,Sucar:etal:ICCV2021}. While classical dense scene representations (e.g. volumetric maps) have several advantages, a trade-off exists between computational cost and topological complexity.  Several papers \cite{Bloesch:etal:CVPR2018,Zhi:etal:CVPR2019,Czarnowski:etal:RAL2020,Sucar:etal:3DV2020} have shown that view-based code representations are able to learn rich prior information from off-line training, enabling joint optimisation of geometry, poses and semantics, to refine network predictions during inference. 3D neural field scene representations have recently gained popularity, owing to their ability to represent complex scene structures with a small memory footprint by exploiting 3D awareness and spatial continuity \cite{Sucar:etal:ICCV2021,Vincent:2019:NIPS2019, Mildenhall:etal:ECCV2020,Park:etal:CVPR2019}. More recently, iMAP \cite{Sucar:etal:ICCV2021} has been proposed as a real-time SLAM system built upon an efficient neural field representation and has demonstrated the ability to reconstruct high-quality, water-tight 3D meshes.

\vspace{-2mm}
\paragraph{Online Scene Understanding and Labelling}
Existing real-time, dense semantic mapping systems typically contain two parallel modules: 1) an RGB-D based geometric SLAM system, maintaining a dense 3D map of the scene, and 2) a semantic segmentation module that predicts dense semantic labels of the scene \cite{Hermans:etal:ICRA2014,Stuckler:Behnke:JVCIR2014,McCormac:etal:3DV2018,Nakajima:etal:ICCV2019}. Multi-view semantic predictions are incrementally fused into the geometric model, yielding densely-labelled, coherent 3D scenes. While semantic segmentation has been performed using a variety of techniques \cite{Nguyen:etal:TVCG2017,Krahenbuhl:Koltun:NIPS2011,Long:etal:CVPR2015,Chen:etal:ECCV2018}, it is an inherently user-dependent and subjective problem \cite{Martin:eta:ICCV2001}. User-in-the-loop systems are therefore crucial in enabling full flexibility when defining semantic relations between entities in a scene. In this context, the works most closely related to ours are SemanticPaint \cite{Valentin:etal:ACMTOG2015} and Semantic Paintbrush \cite{Miksik:etal:CHI2015}. 

SemanticPaint \cite{Valentin:etal:ACMTOG2015} is an online, user-in-the-loop system that allows the user to label a scene during capture. To this end, the user interacts with a 3D volumetric map, built from an RGB-D SLAM system, via voice and hand gestures \cite{Niessner:etal:SIGGRAPH2013}. A streaming random forest classifier, using hand-crafted features, learns continuously from the user gestures in 3D space. The forest predictions are used as unary terms in a conditional random field (CRF) to propagate the user annotations to unseen regions. As the CRFs are built upon the reconstructed data, there is an underlying assumption that this data is good enough to support label propagation. SemanticPaint is therefore restricted to comparably simple scenes and its efficacy in complex real-word scenarios is limited. A significant distinguishing factor between iLabel and SemanticPaint is ease-of-use. SemanticPaint has several distinct modes, requiring the user to switch between modes repeatedly and at well-time intervals to obtain optimal results. In contrast, iLabel offers a much simpler and intuitive user experience, such that high-quality segmentations are obtained with far fewer interactions and no expert knowledge/intuition.

Semantic Paintbrush \cite{Miksik:etal:CHI2015} extends SemanticPaint to operate in outdoor scenes. Using a purely passive stereo setup for extended range and outdoor depth estimation, users visualise the reconstruction through a pair of optical see-through glasses and can draw directly onto it using a laser pointer to annotate the objects in the scene. The system learns in an online manner from the these annotations and is thus able to segment other regions in the 3D map.

In contrast to \cite{Valentin:etal:ACMTOG2015,Miksik:etal:CHI2015}, iLabel does not rely on hand-crafted features, benefiting instead from a powerful joint internal representation of shape and appearance.

\vspace{-4.5mm}
\paragraph{Hierarchical Semantic Segmentation}
Finding the hierarchical structure of complex scenes is a long-standing problem.
Early attempts \cite{Arbalaez:etal:CVPR2014,Arbelaez:etal:TPAMI2010} used image statistics to extract an ultrametric contour map (UCM), leading to further work on using convolutional neural networks (CNNs) for hierarchical image segmentation in a supervised manner \cite{Xie:Tu:ICCV2015,Maninis:etal:ECCV2016,Hiroaki:etal:ARXIV2021}. We show that iLabel can build a user-defined hierarchical scene segmentation interactively and store it within the weights of an MLP.

\vspace{-2mm}
\section{iLabel: Online, Interactive Open-Set Labelling and Learning}
\label{sec:method}
The core real-time SLAM elements of iLabel are similar to iMAP \cite{Sucar:etal:ICCV2021}, which represents 3D scenes using a neural field MLP which maps a 3D coordinate to a colour and volume density. It jointly optimises the MLP and the poses of keyframes through differential volume rendering with actively sampled sparse pixels, while tracking the position of a moving RGB-D camera against the neural representation. 

iLabel adds a semantic head to the MLP that predicts either a flat class distribution or a binary hierarchical tree (see Section \ref{subsec:hierarchical_seg}). While SLAM continues, a user provides annotations via clicks in the keyframes. Scene semantics are then optimised through semantic rendering of these user-selected pixels. The smoothness and compactness priors present in the MLP mean that the user-supplied labels are automatically and densely propagated throughout the scene. iLabel is thus able to produce accurate, dense predictions from very sparse annotations and to often even auto-segment objects and other regions not labelled by the user. The ability to simultaneously reconstruct and label a scene in real-time allows for ultra-efficient labelling of new regions and for easy correction of errors in the current semantic predictions. Figure \ref{fig:overview} gives an overview of the iLabel system.

\subsection{Semantics Representation and Optimisation}
\begin{figure}[htb]
    \centering
    \includegraphics[width =1\linewidth]{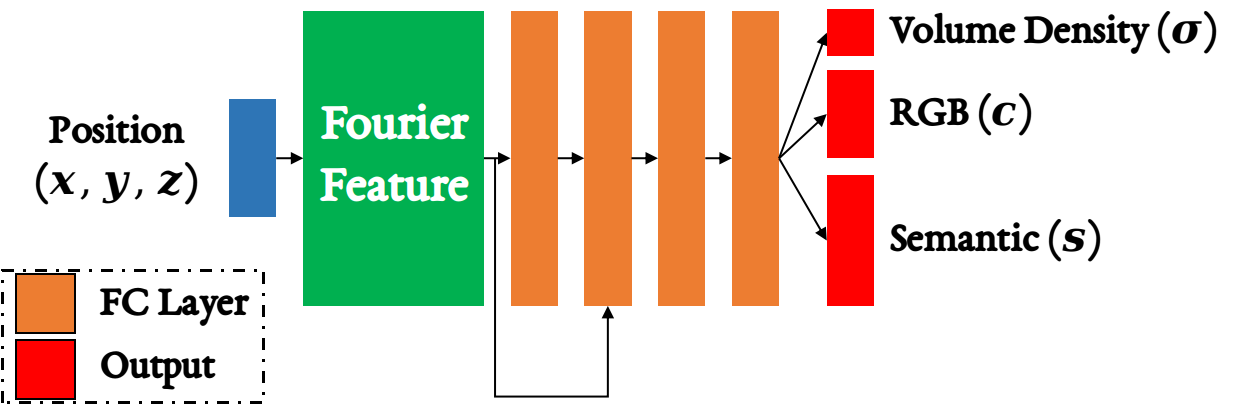}
    \caption{We employ a 4-layer MLP with feature size of 256. }
    \label{fig:MLP_iLabel}
        \vspace{2mm} \hrule
\end{figure}
At the heart of iLabel is continuous optimisation of the underlying implicit scene representation, which follows the network design of iMAP with an additional semantic head (Figure \ref{fig:MLP_iLabel}): 
\begin{equation}{\label{eq:mlp_mapping}}
    F_{\theta}(\textbf{p}) = (\textbf{c},  \textbf{s}, \rho) ,
\end{equation}
where $F_\theta$ is an MLP parameterised by $\theta$; $\textbf{c}$, $\textbf{s}$ and $\rho$ are the radiance, semantic logits and volume density at the 3D position $\textbf{p}=(x,y,z)$, respectively. The scene representation is optimised with respect to volumetric renderings of depth, colour and semantics, computed by compositing the queried network values along the back-projected ray of pixel $[u,v]$:
\begin{equation}{\label{eq:volume_rendering}}
    \small
    \hat{D}[u,v] = \sum_{i=1}^N w_i d_i, \quad \hat{I}[u,v] = \sum_{i=1}^N w_i \textbf{c}_i, \quad \hat{S}[u,v] = \sum_{i=1}^{N} {w_i \textbf{s}_i},
\end{equation}
where $w_i = o_i\prod_{j=1}^{i-1} (1-o_j)$ is the ray-termination probability of sample $i$ at depth $d_i$ along the ray; $o_i = 1 - \exp(-\rho_i\delta_i)$ is the occupancy activation function; $\delta_i = d_{i+1} - d_{i}$ is inter-sample distance.

As in \cite{Sucar:etal:ICCV2021}, geometry and keyframe camera poses $\{T_{WC}\}$ are optimised by minimising the discrepancy between the captured and rendered RGB-D images from sparsely sampled pixels. Semantics are optimised with respect to the user-labelled pixels, with two different activations and losses, corresponding to the two semantic modes described below. The right side of Figure \ref{fig:overview} gives an overview of the semantic rendering process and the activation functions applied to the rendered logits.

\paragraph{Flat Semantics}\label{subsec:flat_seg}

As in \cite{Zhi:etal:ICCV2021}, the network outputs $\textbf{s}_i$ are multi-class semantic logits which are converted into image space by differential volume rendering (Equation \ref{eq:volume_rendering}) followed by a \textit{softmax} activation $\hat{\textbf{S}}[u,v] = \textit{softmax}(\hat{S}[u,v])$. Semantics are then optimised using the image cross-entropy loss between the provided class ID and the rendered predictions.

\paragraph{Hierarchical Semantics}\label{subsec:hierarchical_seg}

We propose a novel hierarchical semantic representation through a binary tree, which allows for labelling and predicting semantics at different hierarchical levels. While the network output, $\textbf{s}_i$, is still represented by an $n$-dimensional flat vector, $n$ now corresponds to the depth of the binary tree as opposed to the number of semantic classes. The semantic logits are rendered in the same manner, but the image activation and loss functions differ.

A \textit{sigmoid} activation function is applied to the rendered logits, producing values in the range $[0,1]$. The $j^{\textup{th}}$ rendered output value, $\hat{\textbf{S}}_j[u,v] = \text{  \textit{sigmoid}}(\hat{S}_j[u,v])$, corresponds to the branching factor at tree level $j$. To obtain a hierarchical semantic prediction, each value $\hat{\textbf{S}}_j[u,v]$ is set to 0 or 1 by thresholding $\hat{\textbf{S}}_j[u,v]$ at 0.5. In the hierarchical setting, the user-supplied label corresponds to selecting a specific node in the binary tree. This label is transformed into a binary branching representation, and a binary cross-entropy loss is computed for each rendered value. A label selecting a tree node at level $L$ only conditions the loss on the output values up to and including level $L$: $\hat{\textbf{S}}_j[u,v], j \in \{ 1,...,L \}$.

With reference to the top half of Figure \ref{fig:hier}, the network outputs three values corresponding to the three levels in the tree. First, the user separates the scene into \textit{foreground} and \textit{background} classes. A background label corresponds to the vector $[0, *, *]$ where $*$ indicates that no loss is calculated for the second and third rendered values. The user then divides the background class further into \textit{wall} and \textit{floor}, where the \textit{wall} label corresponds to vector $[0,1,*]$. The binary hierarchical representation allows the user to separate objects in stages. For example the user first separates a whole bookshelf from the rest of the scene, and later separates the books from the shelf without contradicting the initial labels, meaning that no labelling effort is wasted.

\subsection{Semantic User Interaction Modes}

\begin{figure}[htb]
    \centering
    \includegraphics[width =1\linewidth]{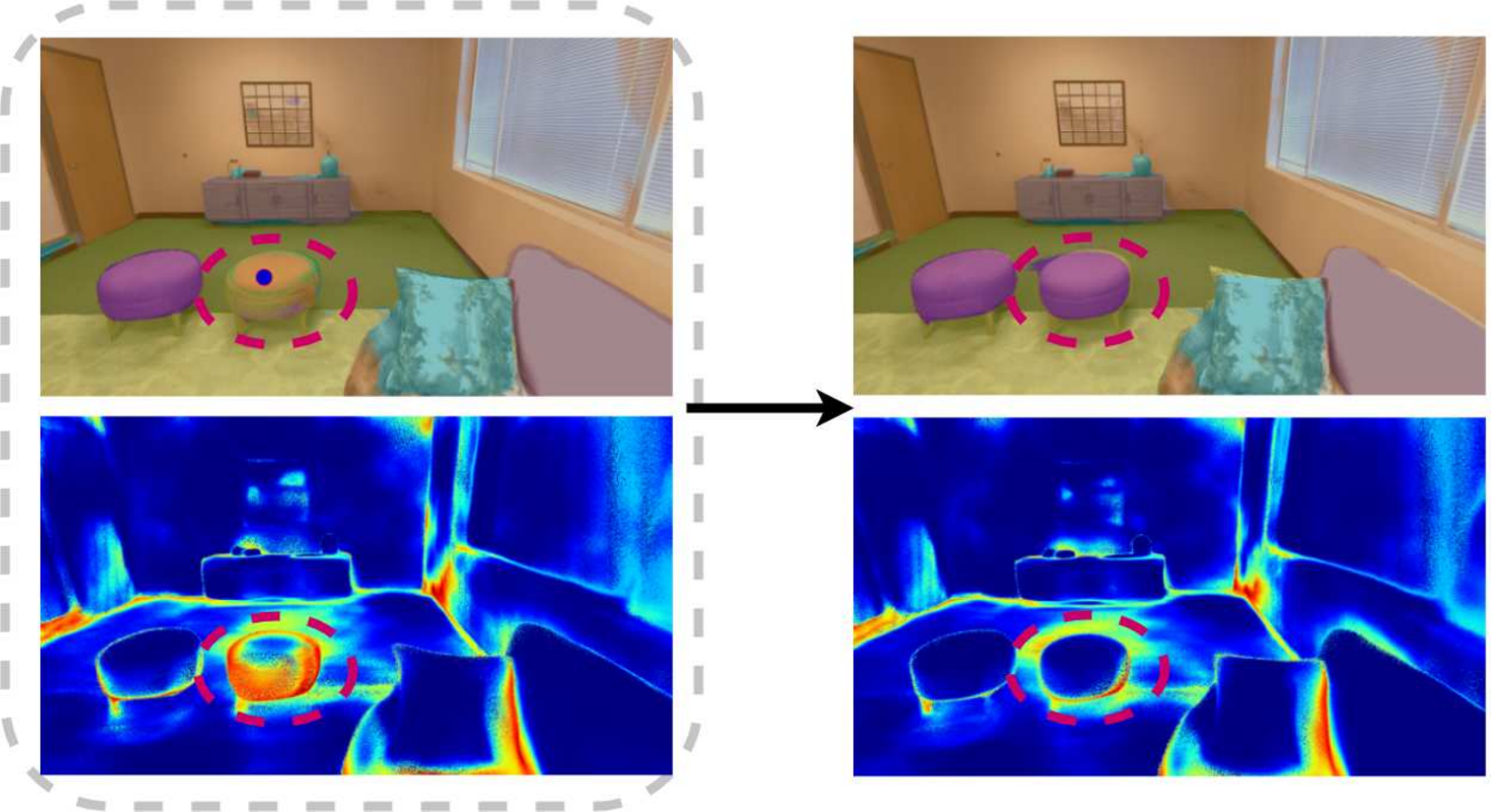}
    \caption{In hands-free mode with automatic query generation, semantic class uncertainty is used to actively select a pixel for which to request a label; in this case an unlabelled stool with ambiguous class prediction and high uncertainty is selected.}
    \label{fig:automate}
        \vspace{2mm} \hrule
\end{figure}

Our system allows for two modes of interaction: 1) \textbf{manual interaction},
the usual interactive mode of iLabel, where users provide semantic labels in image space via clicks, and 2) \textbf{automatic query generation}, where
the system generates automatic queries for the labels of informative pixels, driven by semantic prediction uncertainty (Figure \ref{fig:automate}). The latter mode eases the burden of manual annotation, and users could provide labels via text or voice. 

\vspace{-2mm}
\paragraph{Automatic Query  Generation}
Uncertainty-based sampling is used in this work to actively propose pixel positions for label acquisition because it can integrate seamlessly with deep neural networks with little computational overhead \cite{Settles:Report2009,Ren:etal:ARXIV2020}.
Several uncertainty measures are explored: softmax entropy, least confidence and margin sampling  \cite{Settles:Report2009}. For example, the softmax entropy is defined as:
\begin{equation}\label{eq:entropy}
    u_{entropy} = -\sum_{c=1}^{C}\hat{\textbf{S}}^{c}[u,v] \text{log}(\hat{\textbf{S}}^{c}[u,v]),
\end{equation}
where $C$ is the number of semantic categories. 

At system run-time, semantic labels and corresponding uncertainty maps of all registered keyframes are rendered. 
To decide which keyframe to allocate queries, we first compute frame-level entropy by accumulating pixel-wise entropy within frames and assign a higher probability to sampling the keyframe with higher frame-level entropy. Given a selected keyframe, we then randomly select the queried pixel coordinate from a pool of pixel positions with top-K highest entropy values. The frame-level and pixel-level uncertainty are updated every certainty mapping steps. K is set to $1\%$ or $5\%$ of pixel numbers to avoid repeated queries at nearby positions.

\subsection{Implementation Details}\label{subsec: implementation_details}

iLabel operates in a multiprocessing, single or multi-GPU framework, running three concurrent processes: 1) tracking, 2) mapping, and 3) labelling (see Figure \ref{fig:overview}). 

The mapping process encompasses optimising the MLP parameters with respect to a growing set of $W$ keyframes and associated RGB-D observations: $\{(I_i, D_i, T_i)\}_{i=1}^W$.  As per \cite{Sucar:etal:ICCV2021}, the photometric loss $L_p$ and geometric loss $L_g$ are minimised on sparse, information-guided pixels. iLabel performs an additional optimisation on $K$ user-selected pixels ($\xi_i$) in each keyframe and introduces a semantic loss $L_s$, minimising the following objective function:  
\begin{equation}\label{eq:joint_opt}
    \argmin_{\theta} 
    \frac{1}{K} \sum_{i=1}^W  \sum_{(u,v) \in \xi_i}  \underbrace{e_i^g[u,v]}_{L_g} + \alpha_p \underbrace{e_i^p[u,v]}_{L_p} + \alpha_s \underbrace{e_i^s[u,v]}_{L_s} , 
\end{equation}
where:
{
\footnotesize
\begin{gather*}
    e_i^p[u,v] = \left|I_i[u,v] - \hat{I}_i[u,v]\right|, e_i^s[u,v] = -\sum_{c=1}^{C} \textbf{S}_i^c[u,v]\log (\hat{\textbf{S}}_i^c[u,v]), \\
    e_i^g[u,v] = \frac{\left|D_i[u,v] - \hat{D}_i[u,v]\right|}{\sqrt{\hat{D}_{var}[u, v]}}, \hat{D}_{var}[u,v] = \sum_{i=1}^N w_i (\hat{D}[u,v] - d_i)^2,
\end{gather*}
}
and in the hierarchical setting: 
{\footnotesize
\begin{equation*}
    e_i^s[u,v] = \sum_{l=1}^{L} -\textbf{S}_i^c[u,v]\log (\hat{\textbf{S}}_i^c[u,v]) - (1-\textbf{S}_i^c[u,v])\log (1-\hat{\textbf{S}}_i^c[u,v]).
\end{equation*}}
The labelling process coordinates user interactions (clicks and labels) and controls the rendering of semantic images and meshes (via marching cubes on a dense voxel grid queried from the MLP).
The ADAM optimiser is used with poses and map learning rates of 0.003 and 0.001. $\alpha_c$ and $\alpha_s$ are 5 and 8.

iLabel does not have an explicit/specific refinement process, and all user clicks are involved in the joint optimisation (Equation \ref{eq:joint_opt}). The optimisation keeps working and growing with changing sparse samples for colour and geometry reconstruction, and increasing annotated pixels for semantics, colour and depth as well.

\begin{figure}[tb]
    \centering
    \includegraphics[width =1\linewidth]{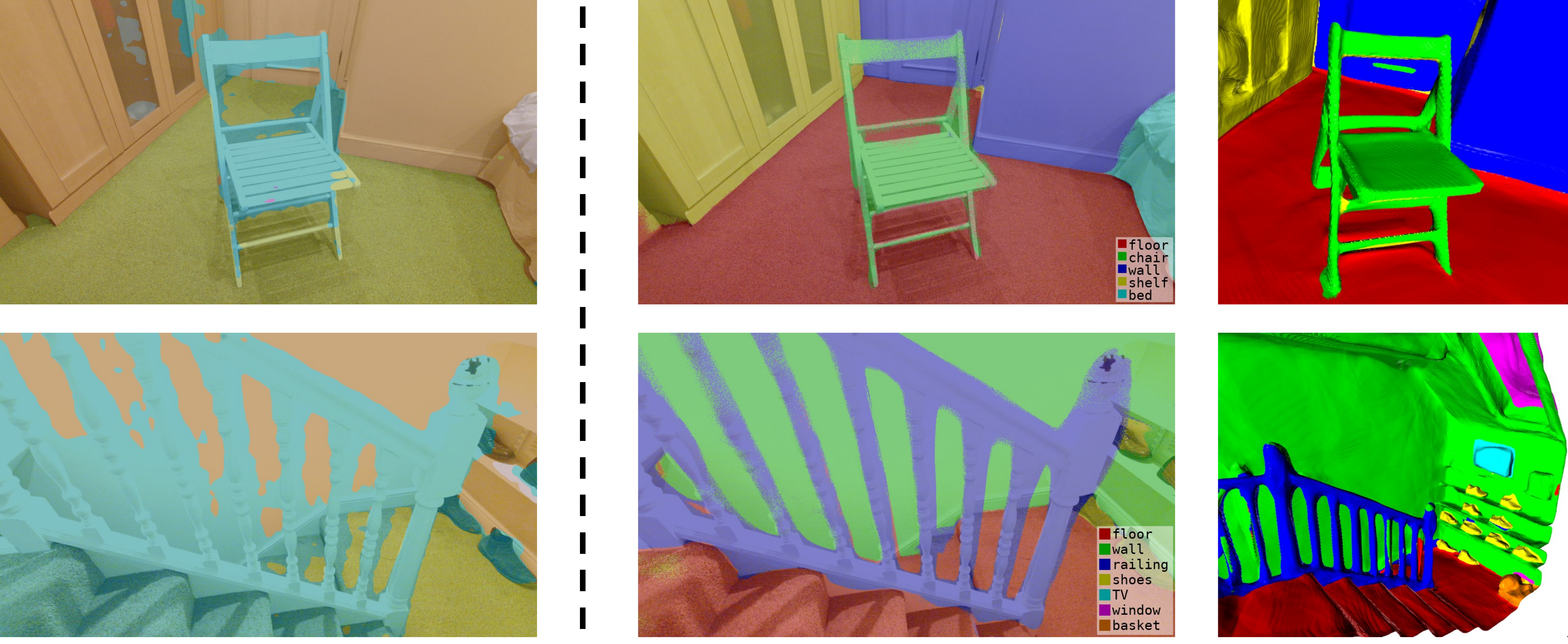}

    \caption{Segmentation results for challenging skeletal objects; left: pre-trained CNN on ScanNet (see Section \ref{subsec:quantitative}), right: iLabel.} 
    \label{fig:thin_objs}
        \vspace{2mm} \hrule
\end{figure}

\begin{figure}[tb]
    \centering
    \includegraphics[width =1\linewidth]{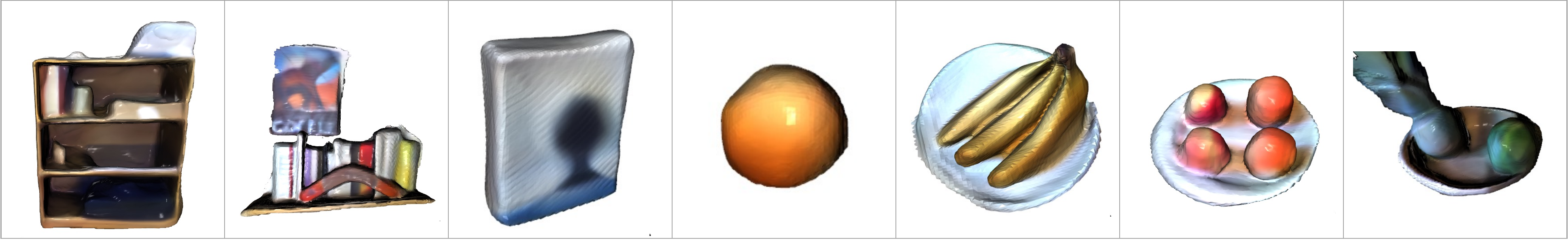}
    \caption{Catalog of object mesh assets separated with iLabel. } 
    \label{fig:catalog}
        \vspace{2mm} \hrule
\end{figure}

\begin{figure}[tb]
    \centering
    \includegraphics[width =1\linewidth]{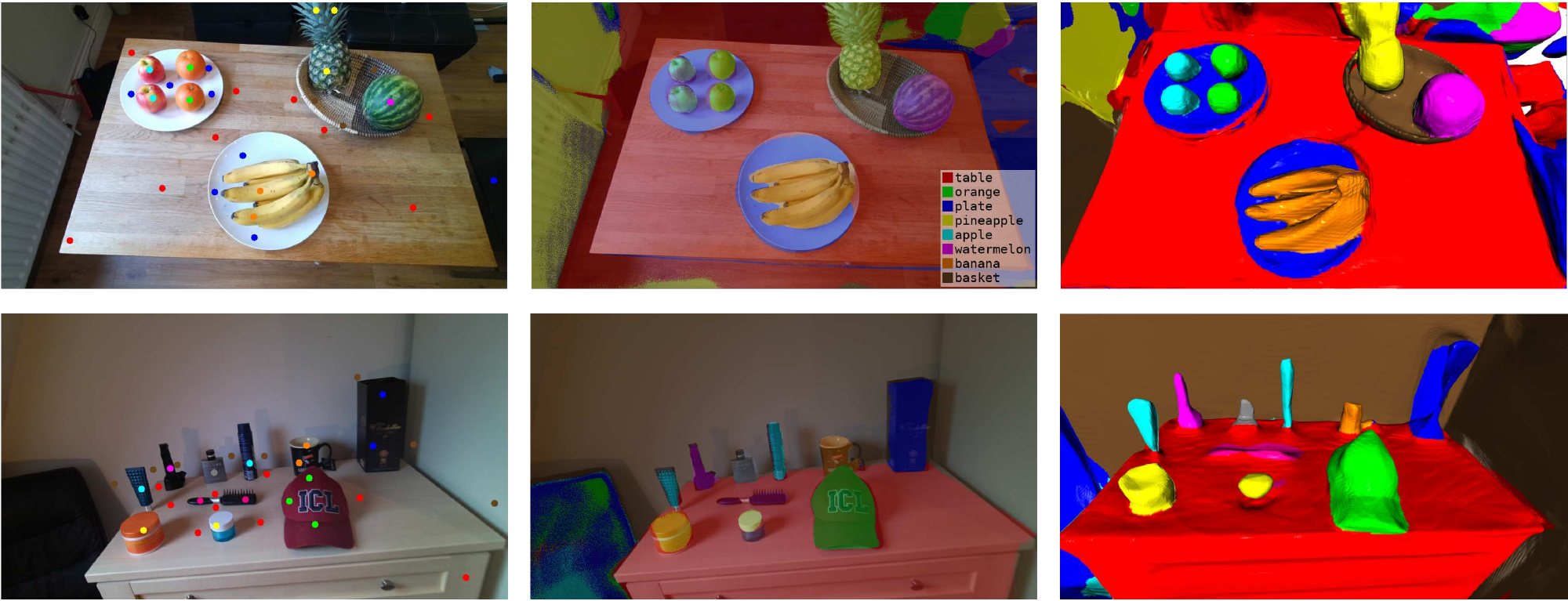}
    \caption{Precise segmentations can be obtained from just 1 or 2 interactive clicks per object. (Left: clicks; middle: dense labels rendered into a keyframe; right: full 3D mesh with labels.)}
    \label{fig:object_seg}    
    \vspace{2mm} \hrule
\end{figure}

\section{Experiments and Applications}
\label{sec:experiments}
\begin{figure*}[tb]
    \centering
    \includegraphics[width =1\linewidth]{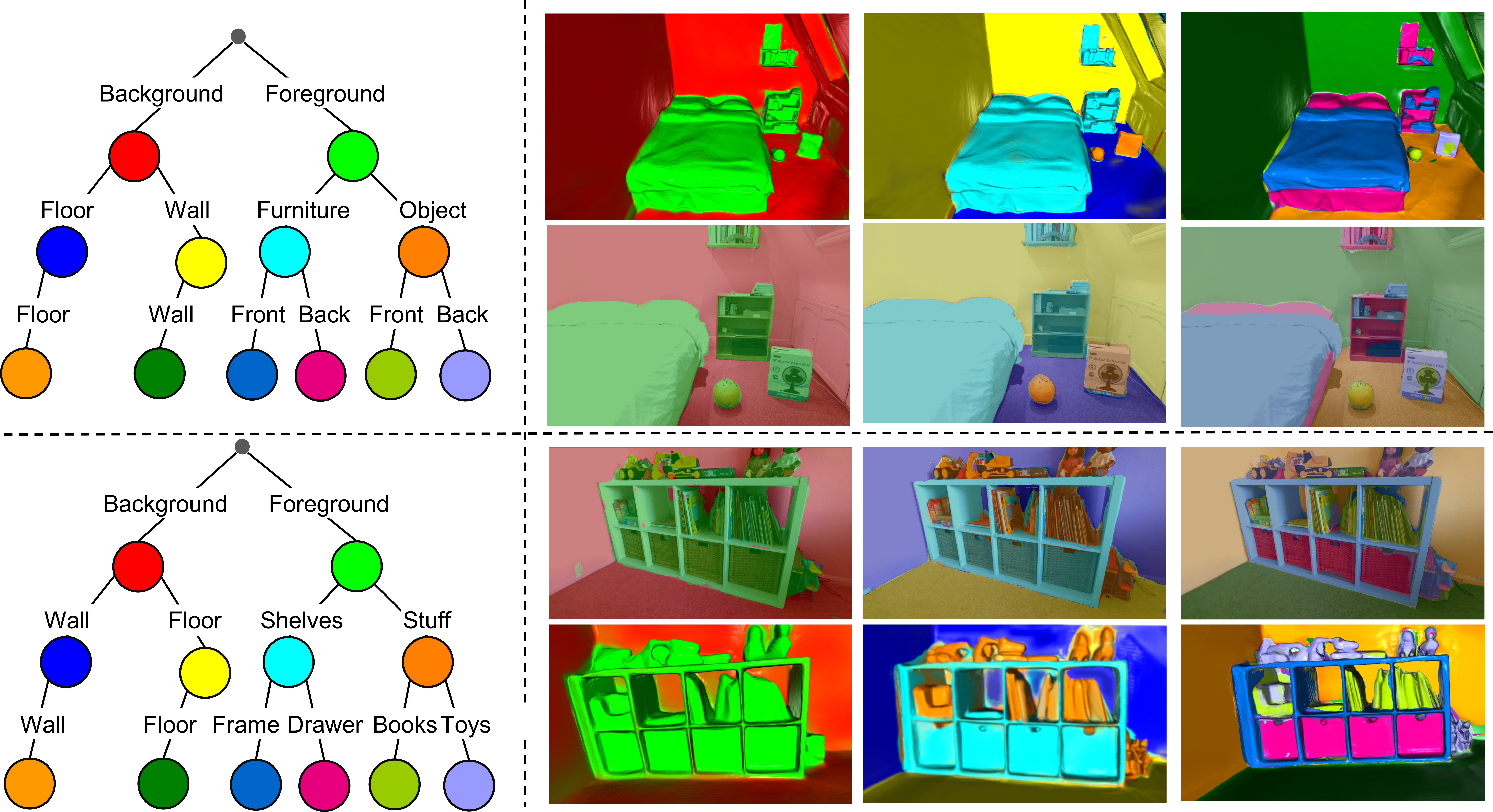}
    \caption{Binary tree as well as the segmentations at each level from the hierarchical mode of iLabel.} 
    \label{fig:hier}
        \vspace{2mm} \hrule
        \vspace{-1mm}
\end{figure*}

iLabel is an interactive tool intended for real-time use and we therefore emphasise that its strengths are best illustrated \textit{qualitatively}. We provide extensive examples to demonstrate iLabel in a variety of interesting scenes, and highly recommend that reviewers watch our {\bf attached video} (Figure \ref{fig:room_seg}) which shows the full interactive labelling process. We show qualitative comparisons with the only comparable system SemanticPaint and clearly demonstrate better segmentation quality. Additionally we perform a quantitative evaluation to show how segmentation quality scales with additional click labels, using  a state-of-the-art, fully-supervised RGB-D segmentation baseline \cite{Chen:etal:ECCV2020}.

\subsection{Qualitative Evaluation}

As the geometry, colour and semantic heads share a single MLP backbone, user annotations are naturally propagated to untouched regions of the scene without specifying an explicit propagation mechanism (e.g. the pairwise terms of a CRF used in \cite{Valentin:etal:ACMTOG2015}). This, together with a user-in-the-loop, enables ultra-efficient scene labelling with only a small number of well-placed clicks.

\begin{figure}[tb]
    \centering
    \includegraphics[width =1\linewidth]{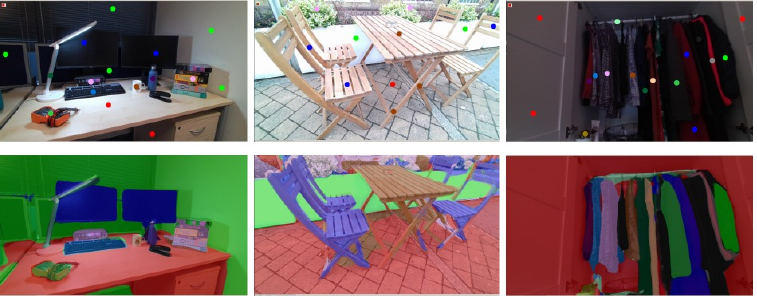}
    \caption{Ultra-efficient label propagation: iLabel produces high-quality segmentations of coherent 3D entities with very few user clicks, approximately 20--30 per scene.} 
    \label{fig:interesting_scenes}
        \vspace{2mm} \hrule
        \vspace{-4mm}
\end{figure}

\begin{figure}[htb]
    \centering
    \includegraphics[width =1\linewidth]{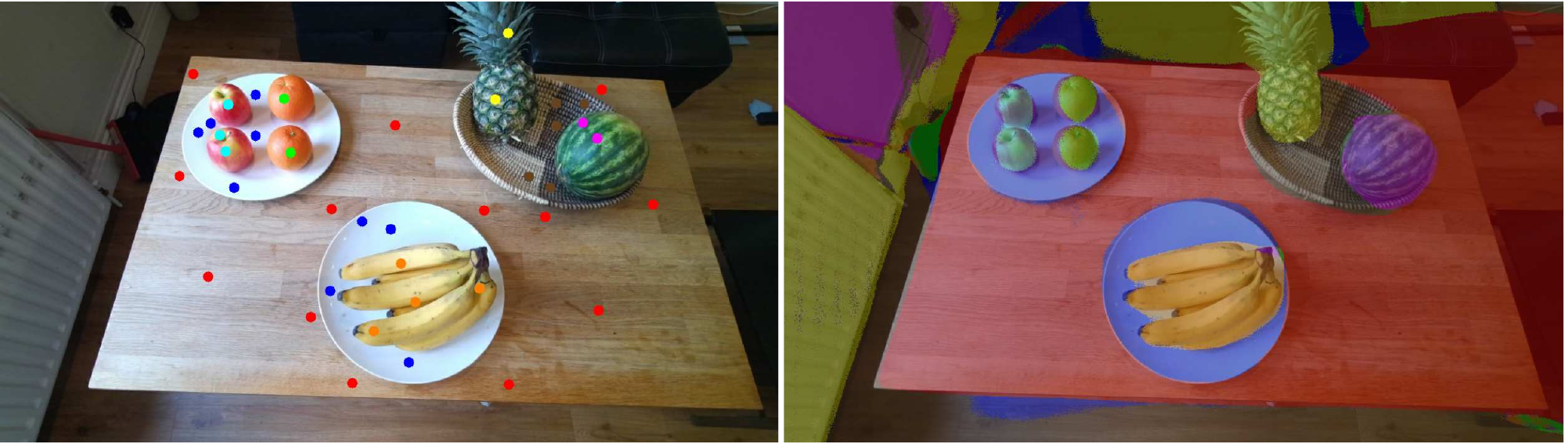}
    \caption{In removing the use of colour optimisation for scene reconstruction, only a few extra clicks are required to achieve a comparable quality of segmentation to that shown in Figure \ref{fig:object_seg}.} 
    \label{fig:no_col}
        \vspace{2mm} \hrule
        \vspace{-2mm}
\end{figure}

 We have observed that the resulting embeddings are highly correlated for coherent 3D entities in the scene (e.g. objects, surfaces, etc.). Consequently, iLabel is able to segment these entities very efficiently, even with a single click. This is illustrated in Figures \ref{fig:object_seg} and \ref{fig:interesting_scenes}, where only a few clicks generate complete and precise segmentations for a wide range of objects and entities, ranging from small, coherent objects (e.g. fruit) to deformable and intricate entities (clothing and furniture). In Figure \ref{fig:no_col} we  disable colour optimisation to further highlight that in iLabel geometry provides a strong signal for separating objects.
 
The coordinate-based representation avoids quantisation and allows the network to be queried at arbitrary resolutions. This property allows reconstruction of detailed geometry and skeletal shapes that, when semantically labelled, render very precise segmentations. Figure \ref{fig:thin_objs} illustrates high-fidelity segmentations of objects which are challenging for a standard CNN. 

iLabel can be used as an efficient tool for generating labelled scene datasets. For example, a scene of a complete room with 13 classes, can be fully segmented with high precision with only 140 user clicks (Figure \ref{fig:room_seg}). Alternatively, iLabel can be used to tag individual objects for generating object-asset catalogues (Figure \ref{fig:catalog}) to aid robotic manipulation tasks, for example. 

\begin{figure}[htb]
    \centering
    \includegraphics[width =1\linewidth]{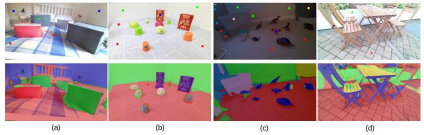}
    \caption{Generalisation: iLabel is able to transfer user labels to objects exhibiting similar properties. It is worth highlighting that the segmentation in (d) was achieved with only 4 clicks.} 
    \label{fig:generalisation}
        \vspace{2mm} \hrule
        \vspace{-3mm}
\end{figure}

While iLabel is particularly powerful at segmenting coherent entities, Figure \ref{fig:generalisation} also demonstrates its ability to propagate user-supplied labels to disjoint objects exhibiting similar properties. Each example shows label transfer between similar objects where only one has been labelled (e.g. (a) boxes on the bed, (b) food boxes and plastic cups and (c) toy dinosaurs). The table and chairs scene in Figure \ref{fig:generalisation} (d) is especially interesting. Only four clicks are supplied: the label for the chair leg (blue) propagates to the leg of the table and the legs of the other chairs, while the table-top label (yellow) propagates to the seats of the chairs.

\begin{figure*}[htbp]
    \centering

    \begin{subfigure}{0.2\linewidth}
        \centering
        \includegraphics[width=0.98\linewidth]{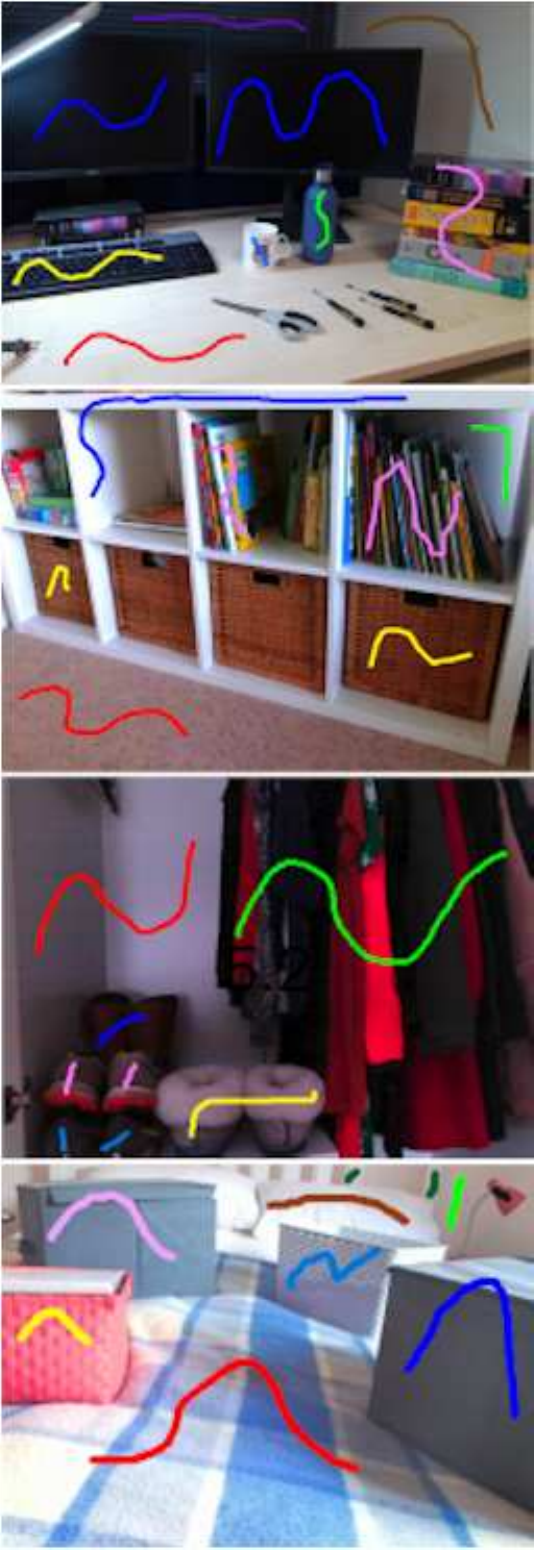}
        \caption{\footnotesize{Input annotations}}
    \end{subfigure}
    \begin{subfigure}{0.2\linewidth}
        \centering
        \includegraphics[width=0.985\linewidth]{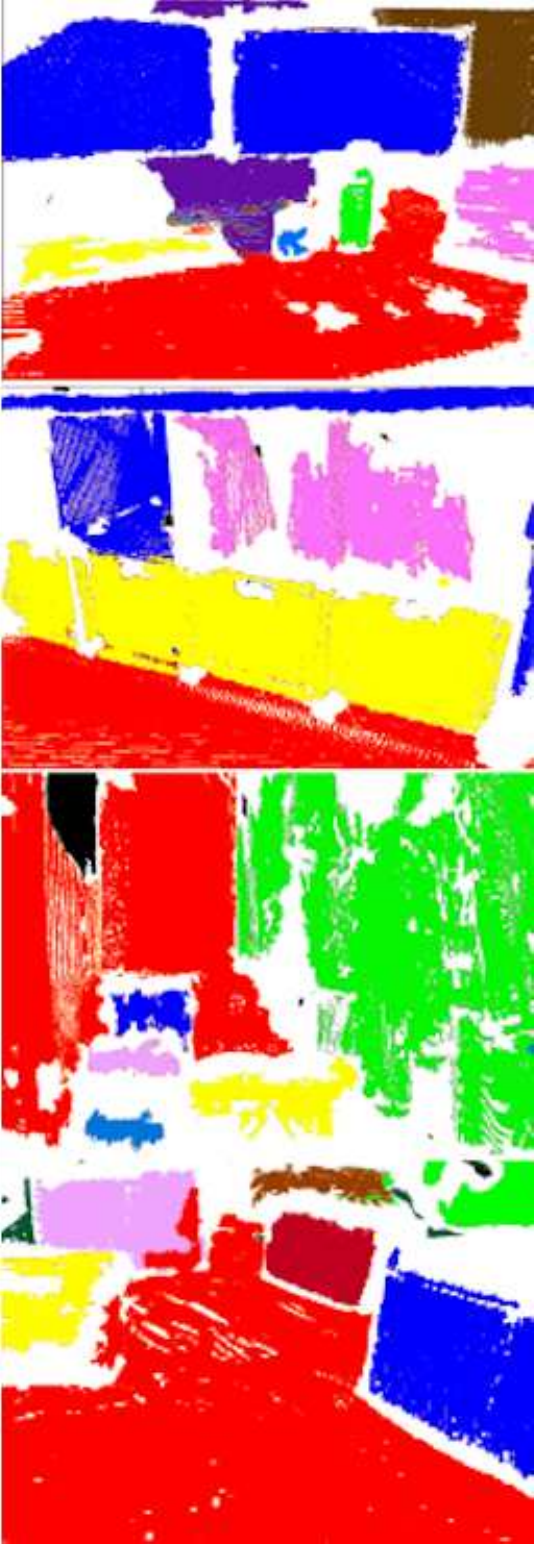}
        \caption{\footnotesize{\textbf{SPaint:} Initial strokes}}
    \end{subfigure}
    \begin{subfigure}{0.2\linewidth}
        \centering
        \includegraphics[width=0.997\linewidth]{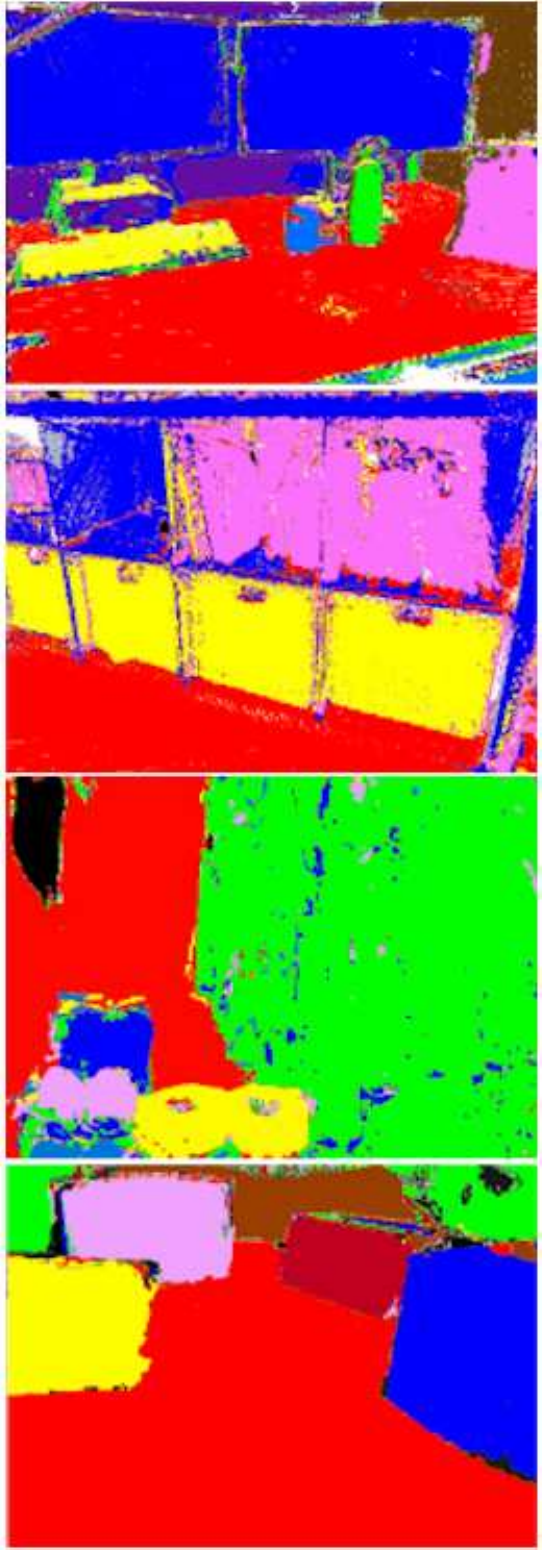}
        \caption{\footnotesize{\textbf{SPaint:} Additional strokes}}
    \end{subfigure}
    \begin{subfigure}{0.2\linewidth}
        \centering
        \includegraphics[width=0.98\linewidth]{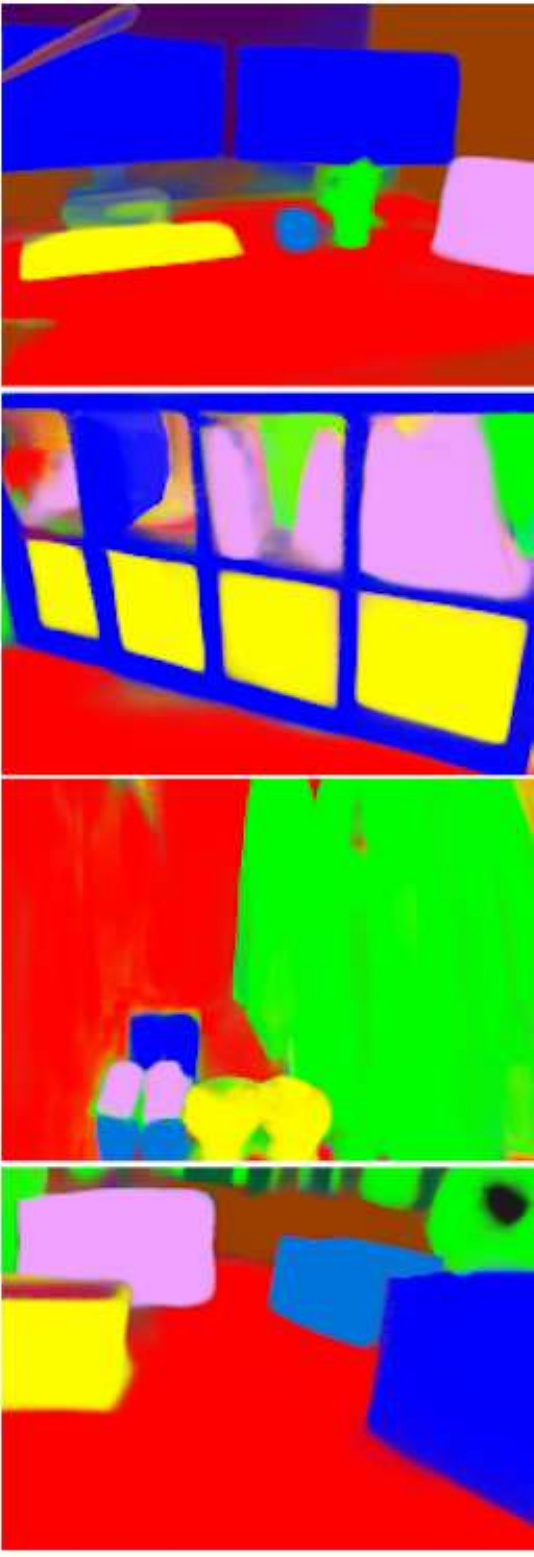}
        \caption{\footnotesize{\textbf{iLabel:} Initial strokes}}
    \end{subfigure}
                
    \caption{Comparison results between iLabel and SemanticPaint for user annotations in (a). (b) SPaint results for initial strokes; (c) SPaint results after corrections; (d) iLabel segmentations obtained using only the input strokes in (a).}
    \label{fig:compare_spaint}
        \vspace{2mm} \hrule
        \vspace{-2mm}
\end{figure*}

\vspace{-4mm}
\paragraph{Hierarchical scene segmentation}
Figure \ref{fig:hier} demonstrates iLabel's hierarchical mode. The colour-coded hierarchy (defined on-the-fly) is shown together with segmentations and scene reconstructions from each level. The results show the capacity of this representation to group objects at different levels, which has potential in applications where different tasks demand different groupings.

\vspace{-4mm}
\paragraph{Comparison to SemanticPaint} \label{sec:compare_spaint}
SemanticPaint (SPaint) \cite{Valentin:etal:ACMTOG2015} is currently the only comparable online interactive scene understanding system. With several distinct modes (labelling, propagation, training, predicting, correcting, smoothing), which do not operate simultaneously, users have to switch between modes repeatedly (with careful consideration given to the duration spent in each mode) to obtain optimal results. In contrast, iLabel presents a unified interface for scene reconstruction, whereby user interaction, label propagation, learning and prediction occur simultaneously. The more intuitive and simpler interface presented by iLabel means that high-quality segmentations are obtained with far fewer interactions and no expert knowledge/intuition.

\begin{figure}[htbp]
    \centering
    \begin{subfigure}{0.32\linewidth}
        \centering
        \includegraphics[width=1\linewidth]{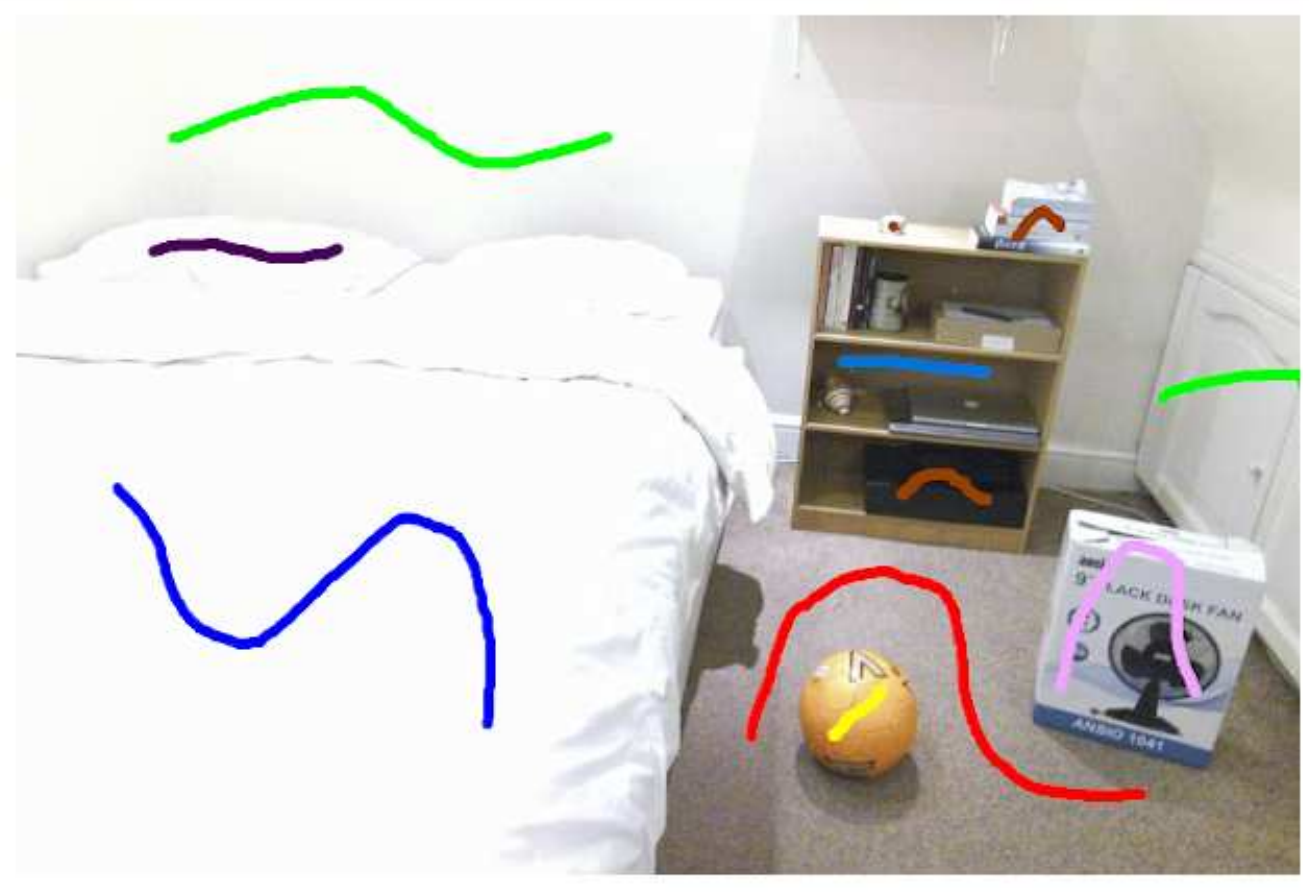}
        \caption{Input annotations}
    \end{subfigure}
    \begin{subfigure}{0.32\linewidth}
        \centering
        \includegraphics[width=1\linewidth]{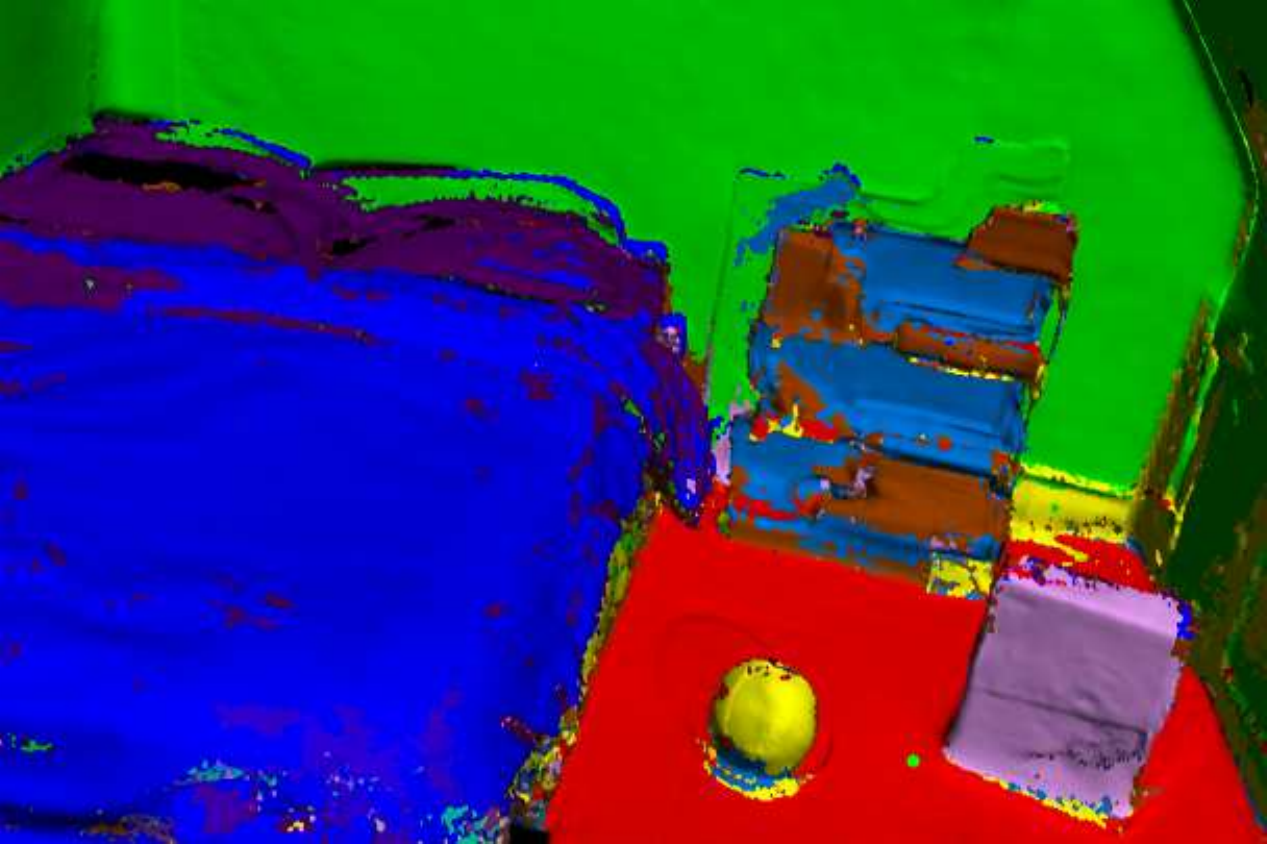}
        \caption{SemanticPaint}
    \end{subfigure}
    \begin{subfigure}{0.32\linewidth}
        \centering
        \includegraphics[width=1\linewidth]{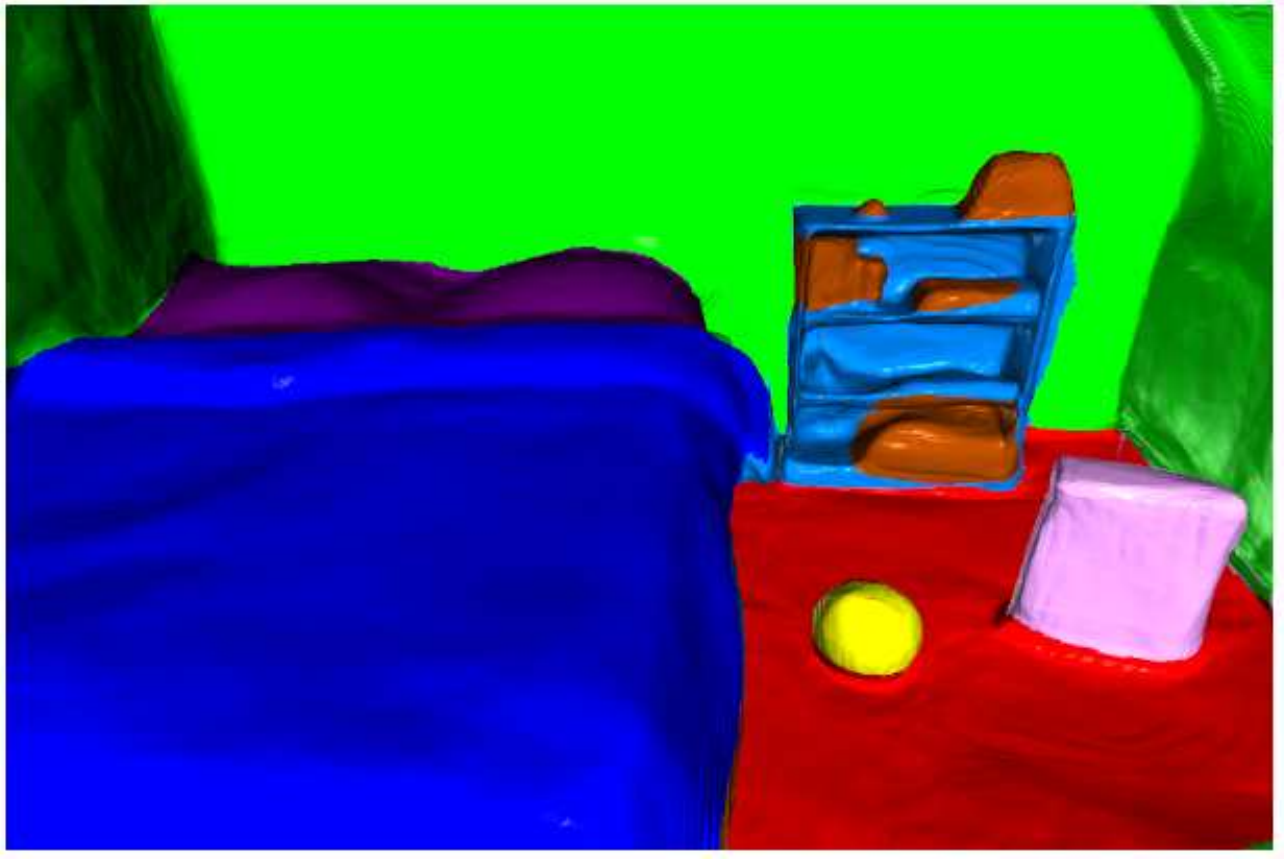}
        \caption{iLabel}
    \end{subfigure}
    \caption{Qualitative comparison between iLabel and SemanticPaint showing generated meshes.} 
    \label{fig:compare_spaint_2}
        \vspace{2mm} \hrule
        \vspace{-2mm}
\end{figure}

Qualitative comparisons between iLabel and SPaint are given in Figures \ref{fig:compare_spaint} and \ref{fig:compare_spaint_2}. Scenes with varying degrees of complexity were chosen to demonstrate the superiority of iLabel even in scenes well-suited to SPaint (e.g. final row Figure \ref{fig:compare_spaint}). For each scene in Figure \ref{fig:compare_spaint}, users annotated objects/regions with the strokes shown in (a). From these initial annotations only, iLabel was able to generate high-quality segmentations (Figure \ref{fig:compare_spaint} (d)). In contrast, SPaint produced comparatively noisy and incomplete initial segmentations (Figure \ref{fig:compare_spaint} (b)). Multiple mode switches and additional corrective strokes were required to generate the final SPaint results (Figure \ref{fig:compare_spaint} (c)). We argue that the results produced by iLabel with only the initial user inputs ($<10$ strokes), surpass those of SPaint after the additional user interactions. Figure \ref{fig:compare_spaint_2} additionally illustrates the quality of the 3D meshes generated by each technique, further highlighting the superiority of iLabel.

\vspace{-2mm}
\subsection{Quantitative evaluation}\label{subsec:quantitative}

We evaluate iLabel's 2D semantic segmentation performance in both user-interaction and automatic query generation modes, with varying numbers of clicks per scene, on the public datasets
Replica \cite{Straub:etal:ARXIV2019} and ScanNet \cite{Dai:etal:CVPR2017}. Both datasets are publicly available for research purposes under their licence. We report the mean Intersection Over Union (mIOU), averaged over ground truth labels remapped to NYU-13 class definitions.

\vspace{-5mm}
\paragraph{Baseline}
While pre-trained segmentation models serve a different purpose than an interactive scene-specific system (to generalise to unseen scenes) we use them as a baseline to demonstrate the labelling efficiency of our system. iLabel scales rapidly with the number of clicks and rapidly surpasses the pretrained model, even when this has been trained on very similar scenes

Performance is evaluated against SA-Gate \cite{Chen:etal:ECCV2020} with a ResNet-101 DeepLabV3+ backbone \cite{Chen:etal:ECCV2018}, which is the current state-of-the-art in RGB-D segmentation. For Replica, we pre-train SA-Gate using the SUN-RGBD dataset \cite{Song:etal:CVPR2015} and fine-tune on our generated Replica sequences to avoid over-fitting. We adopt a leave-one-out strategy, whereby fine-tuning is performed independently for each test scene using the remaining Replica scenes. For ScanNet, we train SA-Gate directly on the official training sets, achieving 63.98\% mIOU on the validation sets of 13 classes. Approximately 11k (9860 and 475 images for our SUN-RGBD training and validation splits, 900 images for Replica fine-tuning) and 25k training images were used for baseline CNN training on each Replica and ScanNet experiment, respectively. The ResNet-101 backbone is initialised with ImageNet pre-trained weights \cite{Russakovsky:etal:ILSVRC15} through all the experiments. As per \cite{Chen:etal:ECCV2020}, depth maps use HHA encoding \cite{Gupta:etal:ECCV2014}, before which fast depth completion \cite{Ku:etal:CRV2018} is used for hole-filling in ScanNet.

\begin{figure} 
\centering
\subfloat[Manual Interaction.\label{subfig:manual}]{\includegraphics[width=0.9\linewidth]{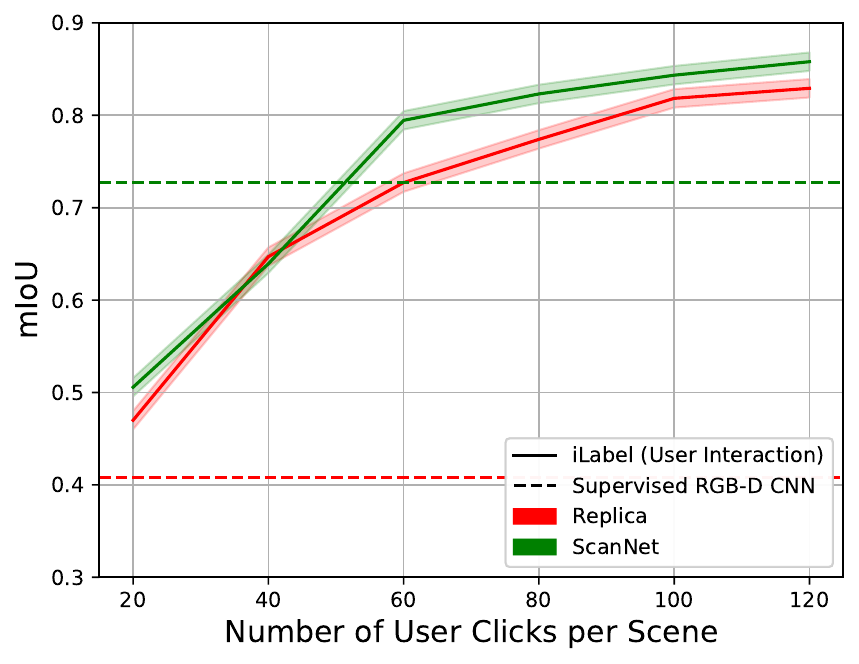}}

\subfloat[Automatic Query Generation.\label{subfig:auto}]{\includegraphics[width=0.9\linewidth]{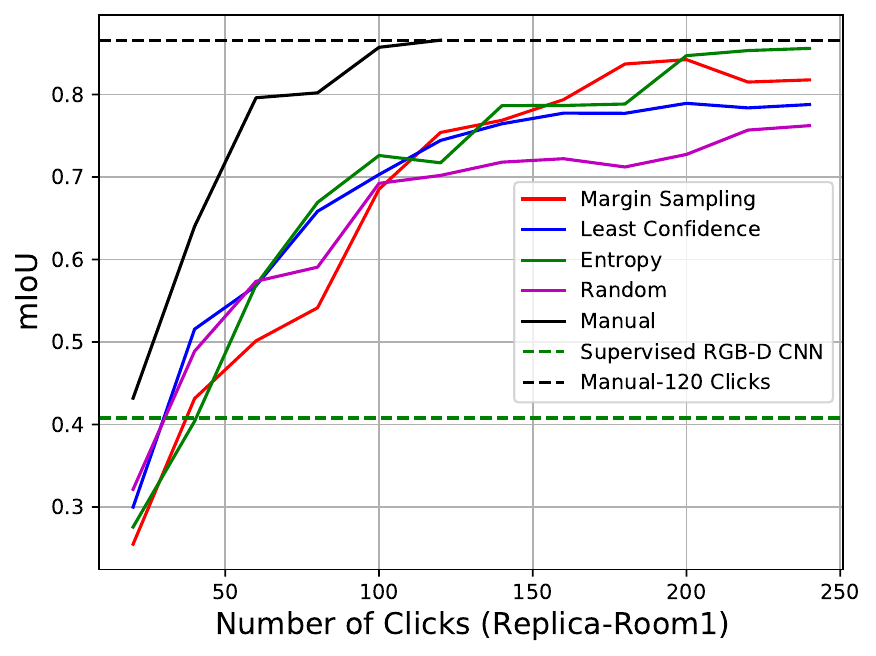}}
\caption{Quantitative evaluation of 2D semantic segmentation on the Replica and ScanNet datasets. Both interaction modes are evaluated and outperform supervised baselines with a small annotation budget.} 
        \vspace{2mm} \hrule
\label{fig:curve}
\vspace{-5mm}
\end{figure}

\vspace{-3mm}
\paragraph{Results}
Figure \ref{subfig:manual} shows the performance of iLabel compared against the supervised RGB-D CNN baseline (dashed horizontal line) on 5 Replica scenes and 6 ScanNet scenes from the validation set. The Replica dataset is a low data regime with only 7 scenes used for fine tuning, which makes generalisation specially hard. iLabel is specially suited for this settings, and surpasses the baseline with only 20 clicks per scene. In the ScanNet dataset where much more data is available, iLabel reaches similar accuracy to the baseline with around 50 clicks, and continues to improve surpassing the baseline by $20\%$ at 120 clicks.

Figure \ref{subfig:auto} shows the effectiveness of automatic query generation, which opens the possibility for hands-free scene labelling, e.g., by voice command. As expected, this mode is less efficient than manual clicks and takes around 240 clicks to reach similar performance. We show how random uniform pixel sampling achieves a lower performance, specially when more labels have been added, highlighting the importance of uncertainty guided pixel selection.

\section{Potential Negative Societal Impacts}
\label{sec:social_impact}
As a visual perception module, iLabel can enable intelligent robots to label novel environments in an open-set manner with only minimal human input. As with any system designed to capture data, user privacy can be negatively impacted. Privacy concerns may be particularly important for iLabel as the scene representations it creates are compact ($\approx$ 1 MB) making the process both portable and scalable. However, these same characteristics may also enable positive technologies such as assistive robotics or inspection platforms that require semantic scene understanding.

\section{Conclusion}
\label{sec:disscusion}
We have shown that online, scene-specific training of a compact MLP model which encodes scene geometry, appearance and semantics allows ultra-sparse interactive labelling to produce accurate dense semantic segmentation, far surpassing the performance of standard pre-trained approaches. Despite promising results, our system's label propagation mechanism works well mainly for proximal regions and/or those sharing similar geometry or texture. A deeper understanding of this mechanism is necessary to enable better control of this process and to improve generalisation performance. 
As  architectures and methods for neural implicit representation of scenes continue to improve, we expect these gains to be passed on to our labelling approach, and for tools like iLabel to become highly practical for applications where users  are able to teach AI systems efficiently about useful scene properties.

\vspace{-3mm}
\section{Acknowledgements}
Research presented in this paper has been supported by Dyson Technology Ltd. Shuaifeng Zhi holds a CSC-Imperial Scholarship.
{\small
\bibliographystyle{ieee_fullname}
\bibliography{robotvision}
}

\end{document}